# A new class of upper bounds on the log partition function

**Martin J. Wainwright**
Stochastic Systems Group,
Dept. of EECS, MIT
Cambridge, MA 02139
mjwain@mit.edu

**Tommi S. Jaakkola**
Lab for Artificial Intelligence,
Dept. of EECS, MIT
Cambridge, MA 02139
tommi@ai.mit.edu

**Alan S. Willsky**
Stochastic Systems Group,
Dept. of EECS, MIT
Cambridge, MA 02139
willsky@mit.edu

### Abstract

Bounds on the log partition function are important in a variety of contexts, including approximate inference, model fitting, decision theory, and large deviations analysis [11, 5, 4]. We introduce a new class of upper bounds on the log partition function, based on convex combinations of distributions in the exponential domain, that is applicable to an arbitrary undirected graphical model. In the special case of convex combinations of tree-structured distributions, we obtain a family of variational problems, similar to the Bethe free energy, but distinguished by the following desirable properties: (i) *they are convex, and have a unique global minimum;* and (ii) *the global minimum gives an upper bound on the log partition function*. The global minimum is defined by stationary conditions very similar to those defining fixed points of belief propagation (BP) or tree-based reparameterization [see 13, 14]. As with BP fixed points, the elements of the minimizing argument can be used as approximations to the marginals of the original model. The analysis described here can be extended to structures of higher treewidth (e.g., hypertrees), thereby making connections with more advanced approximations (e.g., Kikuchi and variants [15, 10]).

## 1 Introduction

A fundamental quantity associated with any graph-structured distribution is the log partition function. For a general undirected model, actually computing the log partition function, though a straightforward summation in principle, is NP-hard due to the exponential number of terms. Therefore, an important problem is either to approximate or obtain bounds on the log partition function. There is a large literature on approximation algorithms for the log partition function [e.g., 7]. A related goal is to obtain upper and lower bounds [e.g., 6, 5]. Such bounds on the log partition function are widely applicable; possible uses include approximate inference [e.g., 5], model fitting [e.g., 4], decision theory, and large deviations analysis [e.g., 11].

An important property of the log partition function is its convexity. Mean field theory [e.g., 4] exploits one aspect of this convexity — namely, that the tangent line is an underestimate [1] — to provide a well-known class of lower bounds on the log partition function. Upper bounds, on the other hand, are not widely available. For the Ising model, Jaakkola and Jordan [6] developed a recursive node-elimination procedure for upper bounding the partition function. This procedure does not appear to have any straightforward generalizations to variables with $m > 2$ states and/or higher order cliques.

In this paper, we exploit exponential representations to derive a new class of upper bounds applicable to an arbitrary undirected graphical model. Bounds in this class are based on taking a particular convex combination of exponential parameter vectors corresponding to distributions in some tractable class.[1] The convex combination is specified by a probability distribution $\vec{\mu}$ over the set of tractable substructures. We consider the problem of optimizing both the choice of exponential parameters, as well as the distribution over tractable subgraphs, so as to obtain the tightest possible bounds. At first sight, this problem appears intractable due to the exponential explosion in its dimensionality. Nonetheless, by a Lagrangian dual reformulation, we obtain a class of variational problems that can be solved efficiently to yield *optimal* upper bounds. In the special case of spanning trees, the con-ditions defining the optima are strikingly similar to the conditions defining fixed points of belief propagation or tree-based reparameterization [13, 14]. However, the dual function has two properties that are not typically enjoyed by the Bethe free energy [15]: it is convex, and the unique global minimum gives an upper bound on the partition function.

---

[1] By tractable, we mean distributions for which inference can be performed efficiently: e.g., graphs of bounded treewidth.

This paper treats the case of discrete random variables, and graphs with pairwise clique potentials; moreover, we assume that the set of tractable substructures, denoted by $\mathfrak{T}$, corresponds to the set of all spanning trees of the graph $G$. Based on an understanding of this case, the modifications necessary to deal with larger clique sizes, and more structured approximations [e.g., 15, 10, 14] will be clear.

This paper is organized in the following manner. We begin in Section 2 by introducing exponential families of distributions. We then define and illustrate the convex combinations of exponential parameters that underlie the basic form of the upper bounds. In Section 3, we derive the optimal form of these upper bounds, by first optimizing over the exponential parameters, and then over the spanning tree distribution. Efficient algorithms for computing upper bounds and their application are presented in Sections 4 and 5 respectively. We conclude in Section 6 with a summary and extensions to this work. A more complete description, including discussion of applications to approximate inference and large deviations analysis, can be found in the thesis [14].

## 2 Notation and set-up

We consider an undirected graph $G = (V, E)$ with $N = |V|$ nodes; in this paper, we assume that the maximal cliques of $G$ have size two. Let $x_s$ be a random variable taking values in the discrete space $\mathcal{X} = \{0, 1, \ldots, m-1\}$, and let $\mathbf{x} = \{x_s \mid s \in V\}$ be a random vector taking values in the Cartesian product space $\mathcal{X}^N$. Our analysis makes use of the following exponential representation of a graph-structured distribution $p(\mathbf{x})$. For some index set $\mathcal{I}$, we let $\phi = \{\phi_\alpha \mid \alpha \in \mathcal{I}\}$ denote a collection of potential functions defined on the cliques of $G$, and let $\theta = \{\theta_\alpha \mid \alpha \in \mathcal{I}\}$ be a vector of weights on these potential functions. The exponential family determined by $\phi$ is the following collection of Gibbs distributions:

$$p(\mathbf{x}; \theta) = \exp\left\{\sum_\alpha \theta_\alpha \phi_\alpha(\mathbf{x}) - \Phi(\theta)\right\} \quad (1a)$$

$$\Phi(\theta) = \log \sum_{\mathbf{x} \in \mathcal{X}^N} \exp\left\{\sum_\alpha \theta_\alpha \phi_\alpha(\mathbf{x})\right\} \quad (1b)$$

where $\Phi(\theta)$ is the *log partition function* that serves to normalize the distribution.

The following well-known properties of $\Phi$ are critical to our analysis:

**Lemma 1.** *(a) For all indices $\alpha \in \mathcal{I}$, we have*

$$\frac{\partial \Phi(\theta)}{\partial \theta_\alpha} = \mathbb{E}_\theta[\phi_\alpha] = \sum_{\mathbf{x} \in \mathcal{X}^N} p(\mathbf{x}; \theta) \phi_\alpha(\mathbf{x})$$

*(b) Moreover, the second derivative is given by an element of the Fisher information matrix — namely:*

$$\frac{\partial^2 \Phi(\theta)}{\partial \theta_\alpha \partial \theta_\beta} = \mathbb{E}_\theta[\phi_\alpha \phi_\beta] - \mathbb{E}_\theta[\phi_\alpha] \mathbb{E}_\theta[\phi_\beta]$$

*so that the log partition function $\Phi$ is convex as a function of $\theta$.*

In a *minimal* representation, the functions $\{\phi_\alpha\}$ are linearly independent. For example, one minimal representation of a binary-valued random vector on a graph with pairwise cliques is the usual Ising model, in which $\phi = \{x_s \mid s \in V\} \cup \{x_s x_t \mid (s,t) \in E\}$. Here the index set $\mathcal{I} = V \cup E$.

In most of our analysis (other than Examples 1 and 2, and our simulations), we use an *overcomplete* representation, in which there are linear dependencies among the potentials $\{\phi_\alpha\}$. In particular, given an $m$-state process ($\mathcal{X} = \{0, 1, \ldots, m-1\}$), we use indicator functions as potentials:

$$\phi_{s;j}(x_s) = \delta(x_s = j), \quad s \in V; \; j \in \mathcal{X}$$
$$\phi_{st;jk}(x_s, x_t) = \delta(x_s = j, x_t = k), \; (s,t) \in E; \; j, k \in \mathcal{X}$$

In this case, the index set $\mathcal{I}$ consists of the union of $\mathcal{I}(V) = \{(s;j) \mid s \in V; \; j \in \mathcal{X}\}$ with the edge indices $\mathcal{I}(E) = \{(st;jk) \mid (s,t) \in E; \; j,k \in \mathcal{X}\}$.

### 2.1 Convex combinations

Let $\mathfrak{T} = \mathfrak{T}(G)$ denote the set of all spanning trees of $G$. For each spanning tree $\mathcal{T} \in \mathfrak{T}$, let $\theta(\mathcal{T})$ be an exponential parameter vector of the same dimension as $\theta$ that respects the structure of $\mathcal{T}$. To be explicit, if $\mathcal{T}$ is defined by an edge set $E(\mathcal{T}) \subset E$, then $\theta(\mathcal{T})$ must have zeros in all elements corresponding to edges not in $E(\mathcal{T})$. For compactness in notation, let $\boldsymbol{\theta} \triangleq \{\theta(\mathcal{T}) \mid \mathcal{T} \in \mathfrak{T}\}$ denote the full collection of tree-structured exponential parameter vectors. The notation $\theta(\mathcal{T})$ specifies those subelements of $\boldsymbol{\theta}$ corresponding to spanning tree $\mathcal{T}$.

In order to define a convex combination, we require a probability distribution $\vec{\mu}$ over the set of spanning trees — that is, a collection of non-negative numbers

$$\vec{\mu} \triangleq \{\mu(\mathcal{T}), \; \mathcal{T} \in \mathfrak{T} \mid \mu(\mathcal{T}) \geq 0\} \quad (2)$$

such that $\sum_{\mathcal{T} \in \mathfrak{T}} \mu(\mathcal{T}) = 1$. For any distribution $\vec{\mu}$, we define its *support*, denoted $\mathrm{supp}(\vec{\mu})$, to be the set of trees to which it assigns non-zero probability. In the sequel, we will also be interested in the probability $\mu_e = \Pr_{\vec{\mu}}\{e \in \mathcal{T}\}$ that a given edge $e \in E$ appears in a spanning tree $\mathcal{T}$ chosen randomly under $\vec{\mu}$. We let $\boldsymbol{\mu_e} = \{\mu_e \mid e \in E\}$ represent a vector of these *edge appearance probabilities*. It can be shown [14] that these edge appearance vectors must belong to the so-called *spanning tree polytope*, denoted by $\mathbb{T}(G)$.

A *convex combination* of exponential parameter vectors is defined via the weighted sum $\sum_{\mathcal{T} \in \mathfrak{T}} \mu(\mathcal{T})\theta(\mathcal{T})$, which we denote compactly as $\mathbb{E}_{\vec{\mu}}[\theta(\mathcal{T})]$. Let $\theta^*$ denote the exponential parameter vector of the distribution $p(\mathbf{x}; \theta^*)$ of interest, which we assume to be intractable. Then we are especially interested in collections of exponential parameters $\boldsymbol{\theta}$ for which there exists a convex combination that is equal to $\theta^*$. Accordingly, we define the set $\mathcal{A}(\theta^*) \triangleq \{(\boldsymbol{\theta}; \vec{\mu}) \mid \mathbb{E}_{\vec{\mu}}[\theta(\mathcal{T})] = \theta^*\}$. It is not difficult to see that $\mathcal{A}(\theta^*)$ is never empty.

**Example 1 (Single cycle graph).** As an illustration of these definitions, consider a binary distribution defined by a single cycle on 4 nodes. Consider a target distribution in the minimal Ising form $p(\mathbf{x}; \theta^*) = \exp\{x_1 x_2 + x_2 x_3 + x_3 x_4 + x_4 x_1 - \Phi(\theta^*)\}$. That is, the target distribution is specified by the minimal parameter $\theta^* = [0\ 0\ 0\ 0\ 1\ 1\ 1\ 1]$, where the zeros represent the fact that $\theta^*_s = 0$ for all $s \in V$. The tractable class consists of the four possible spanning trees $\mathfrak{T} = \{\mathcal{T}_i \mid i = 1, \ldots, 4\}$ on a single cycle on four nodes, as shown in Figure 1. We define a set of associated exponential parameters $\boldsymbol{\theta} = \{\theta(\mathcal{T}_i)\}$ as follows:

$$
\begin{aligned}
\theta(\mathcal{T}_1) &= (4/3)\,[0\ 0\ 0\ 0\ 1\ 1\ 1\ 0] \\
\theta(\mathcal{T}_2) &= (4/3)\,[0\ 0\ 0\ 0\ 1\ 1\ 0\ 1] \\
\theta(\mathcal{T}_3) &= (4/3)\,[0\ 0\ 0\ 0\ 1\ 0\ 1\ 1] \\
\theta(\mathcal{T}_4) &= (4/3)\,[0\ 0\ 0\ 0\ 0\ 1\ 1\ 1]
\end{aligned}
$$

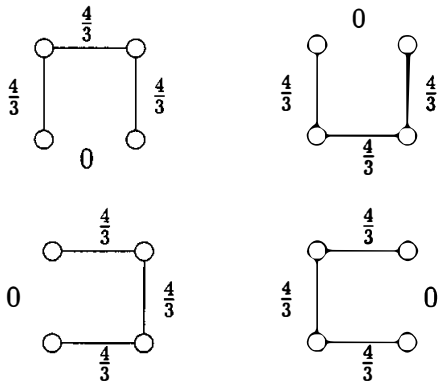

**Figure 1.** A convex combination of four distributions $p(\mathbf{x}; \theta(\mathcal{T}_i))$, each defined by a spanning tree $\mathcal{T}_i$, is used to approximate the target distribution $p(\mathbf{x}; \theta^*)$ on the single-cycle graph.

Finally, we choose $\mu(\mathcal{T}_i) = 1/4$ for all $\mathcal{T}_i \in \mathfrak{T}$. With this uniform distribution over trees, we have $\mu_e = 3/4$ for each edge, and moreover, $\mathbb{E}_{\vec{\mu}}[\theta(\mathcal{T})] = \theta^*$. That is, the specified pair $(\boldsymbol{\theta}; \vec{\mu})$ belongs to $\mathcal{A}(\theta^*)$.

## 2.2 Basic form of upper bound

The convexity of $\Phi$ (see Lemma 1) allows us to apply Jensen's inequality to a convex combination specified by a pair $(\boldsymbol{\theta}, \vec{\mu}) \in \mathcal{A}(\theta^*)$, thereby yielding the upper bound:

$$\Phi(\theta^*) \leq \mathbb{E}_{\vec{\mu}}[\Phi(\theta(\mathcal{T}))] \triangleq \sum_{\mathcal{T} \in \mathfrak{T}} \mu(\mathcal{T})\Phi(\theta(\mathcal{T})) \quad (3)$$

Note that the bound of equation (3) is a function of both the distribution $\vec{\mu}$ over spanning trees; and the collection $\boldsymbol{\theta}$ of tree-structured exponential parameter vectors. In this paper, we shall consider the problem of optimizing these choices so as to minimize the RHS of equation (3), thereby obtaining the tightest possible upper bound. Despite the relatively simple form of equation (3), these optimization problems turn out to have a rich and interesting structure.

## 3 Optimal forms of upper bounds

In this section, we consider first the problem of optimizing the choice of $\boldsymbol{\theta}$ for a fixed $\vec{\mu}$; and then the joint optimization of $\boldsymbol{\theta}$ and $\vec{\mu}$.

### 3.1 Optimizing with $\vec{\mu}$ fixed

For a fixed distribution $\vec{\mu}$, consider the following constrained optimization problem:

$$\begin{cases} \min_{\boldsymbol{\theta}} \mathbb{E}_{\vec{\mu}}[\Phi(\theta(\mathcal{T}))] \\ \text{s.t} \quad \mathbb{E}_{\vec{\mu}}[\theta(\mathcal{T})] = \theta^* \end{cases} \quad (4)$$

With $\vec{\mu}$ fixed, the upper bound $\mathbb{E}_{\vec{\mu}}[\Phi(\theta(\mathcal{T}))]$ is convex as a function of $\boldsymbol{\theta} = \{\theta(\mathcal{T}) \mid \mathcal{T} \in \mathfrak{T}\}$, and the associated constraint is linear in $\boldsymbol{\theta}$.

We assume that $\vec{\mu}$ is chosen such that the associated edge appearance probabilities $\mu_e = \Pr_{\vec{\mu}}\{e \in \mathcal{T}\}$ are all strictly positive. I.e., all edges $e \in E$ appear in at least one tree $\mathcal{T} \in \text{supp}(\vec{\mu})$. This assumption is necessary to ensure that constraint set $\{\boldsymbol{\theta} \mid (\boldsymbol{\theta}, \vec{\mu}) \in \mathcal{A}(\theta^*)\}$ is non-empty. By standard results in nonlinear programming [1], problem (4) has a global minimum, attained at $\widehat{\boldsymbol{\theta}} \equiv \widehat{\boldsymbol{\theta}}(\vec{\mu})$; moreover, it could be solved, in principle, by a variety of methods. However, an obvious concern is the dimension of the parameter vector $\boldsymbol{\theta}$: it is directly proportional to $|\mathfrak{T}|$, the number of spanning trees in $G$, which is typically very large.[2]

---
[2] For example, the complete graph on $N$ nodes has $N^{N-2}$ spanning trees [2].

However, the theory of convex duality allows us to neatly avoid this combinatorial explosion. In particular, we show that the Lagrangian dual of problem (4) depends on a vector of *pseudomarginals* on the nodes and edges of the graph:

$$\mathbf{T} = \{T_s,\ s \in V\ \} \cup \{\ T_{st},\ (s,t) \in E\ \} \quad (5)$$

The constraint set $\mathbb{C}$ for $\mathbf{T}$ consists of the local consistency conditions:

$$\mathbb{C} \triangleq \{\ \mathbf{T}\ |\ \sum_{x'_t} T_{st}(x_s, x'_t) = T_s(x_s),\ \sum_{x'_s} T_s(x'_s) = 1\ \}$$

Let $\widehat{\theta} = \{\ \widehat{\theta}(\mathcal{T})\ |\ \mathcal{T} \in \mathfrak{T}\ \}$ denote the optimum of problem (4). The significance of $\mathbf{T}$ is in specifying this optimum in a very compact fashion. For each tree $\mathcal{T} \in \mathfrak{T}$, let $\Pi^{\mathcal{T}}(\mathbf{T})$ denote the projection of $\mathbf{T}$ onto the spanning tree $\mathcal{T}$. Explicitly,

$$\Pi^{\mathcal{T}}(\mathbf{T}) \triangleq \{T_s,\ s \in V\ \} \cup \{\ T_{st},\ (s,t) \in E(\mathcal{T})\ \} \quad (6)$$

consists only of those elements of $\mathbf{T}$ corresponding to single nodes, or belonging to the edge set $E(\mathcal{T}) \subset E$ of the tree $\mathcal{T}$. Any such vector $\Pi^{\mathcal{T}}(\mathbf{T})$ provides an explicit construction of a distribution $p(\mathbf{x}; \Pi^{\mathcal{T}}(\mathbf{T}))$ via the usual factorization of tree-structured distributions implied by the junction tree representation [9] — viz.:

$$p(\mathbf{x}; \Pi^{\mathcal{T}}(\mathbf{T})) \triangleq \prod_{s \in V} T_s(x_s) \prod_{(s,t) \in E(\mathcal{T})} \frac{T_{st}(x_s, x_t)}{T_s(x_s)\, T_t(x_t)}$$

The proof of Theorem 1 below shows that the optimal dual parameter $\widehat{\mathbf{T}}$ specifies the full collection of optimal exponential parameters $\widehat{\theta}$ via the relation:

$$p(\mathbf{x}; \widehat{\theta}(\mathcal{T})) \propto p(\mathbf{x}; \Pi^{\mathcal{T}}(\widehat{\mathbf{T}})) \quad \text{for all } \mathcal{T} \in \mathfrak{T} \quad (7)$$

That is, at the optimum, a single collection of pseudomarginals $\mathbf{T}$ on nodes and edges suffices to specify the full collection $\widehat{\theta} = \{\ \widehat{\theta}(\mathcal{T})\ |\ \mathcal{T} \in \mathfrak{T}\ \}$ of tree parameters. Consequently, the dual formulation reduces the problem dimension from the size of $\theta$, which is proportional to $|\mathfrak{T}|$, down to the dimension of $\mathbf{T}$ — namely, $(mN + m^2|E|)$. It is this massive reduction in the problem dimension that permits efficient computation of the optimum. Moreover, the conditions in equation (7) that define the optimal $\widehat{\mathbf{T}}$ are very similar to the consistency conditions satisfied by any fixed point of tree-based reparameterization or BP [13, 14]. Not surprisingly then, the dual function of Theorem 1 has a very close relation with the Bethe free energy.

With this intuition, we now state and prove the optimal upper bounds of the type in equation (3). Let $\vec{\mu}$ be a distribution over spanning trees, with associated edge appearance probabilities $\mu_e$. For each $s \in V$, let $H_s(T_s)$ denote the entropy of the distribution $T_s$; for each $(s,t) \in E$, let $I_{st}(T_{st})$ denote the mutual information between $x_s$ and $x_t$ as measured under $T_{st}$. Lastly, let $\sum_{\alpha \in \mathcal{I}} T_\alpha \theta^*_\alpha$ be a compact representation of the two terms in the "average energy" — namely $\sum_{s \in V} \sum_{j \in \mathcal{X}} T_s(x_s = j)\theta^*_{s;j}$ and $\sum_{(s,t) \in E} \sum_{j,k \in \mathcal{X}} T_{st}(x_s = j, x_t = k)\theta^*_{st;jk}$. Using this notation, our bounds are based on the following function:

$$\mathcal{F}(\mathbf{T}; \mu_e; \theta^*) \triangleq -\sum_{s \in V} H_s(T_s) + \sum_{(s,t) \in E} \mu_{st} I_{st}(T_{st}) - \sum_{\alpha \in \mathcal{I}} T_\alpha \theta^*_\alpha \quad (8)$$

**Theorem 1 (Optimal upper bounds).**
*For an arbitrary $\vec{\mu} \in \mathbb{T}(G)$, $\mathcal{F}(\mathbf{T}; \mu_e; \theta^*)$ is convex as a function of $\mathbf{T}$. Moreover, the log partition function is bounded above as follows:*

$$\Phi(\theta^*) \leq -\min_{\mathbf{T} \in \mathbb{C}} \left\{ \mathcal{F}(\mathbf{T}; \mu_e; \theta^*) \right\} \quad (9)$$

*and this global minimum is attained at a unique vector $\widehat{\mathbf{T}} \equiv \widehat{\mathbf{T}}(\mu_e)$ in the constraint set $\mathbb{C}$.*

**Remarks:** (1) Note that when $\mu_e = 1$, then $\mathcal{F}(\mathbf{T}; 1; \theta^*)$ is equivalent to the Bethe free energy [15]. Of course, each edge can belong to every spanning tree with probability one (i.e., $\mu_e = 1$ for all $e \in E$) *only* when the graph $G$ is actually a tree.

(2) Equation (9) stipulates the bound that is optimal over all bounds of the form in equation (3). The vector $\widehat{\mathbf{T}}$ specifies the optimal collection of tree-structured exponential parameters $\widehat{\theta} = \{\ \widehat{\theta}(\mathcal{T})\ \}$ in a very compact fashion via equation (7).

(3) The convexity (and the resultant unique global minimum) of the variational problem in Theorem 1 is in sharp contrast with mean field theory, where the associated optimization problem is well-known to suffer from multiple local minima, even for relatively simple problems [e.g., 4].

*Proof of Theorem 1:* We shall calculate the Lagrangian dual of problem (4), and show that it is equivalent to $-\mathcal{F}$. We form the Lagrangian:

$$\mathcal{L}(\theta; \mathbf{T}; \vec{\mu}; \theta^*) = \mathbb{E}_{\vec{\mu}}[\Phi(\theta(\mathcal{T})] + \sum_{\alpha \in \mathcal{I}} T_\alpha \{\theta^*_\alpha - \mathbb{E}_{\vec{\mu}}[\theta(\mathcal{T})_\alpha]\}$$

where the vector $\mathbf{T} = \{T_s, T_{st}\}$ corresponds to Lagrange multipliers.[3] (E.g., the quantity $T_{s;j}$ is associated with the constraint that $\mathbb{E}_{\vec{\mu}}[\theta_{s;j}(\mathcal{T})] = \theta^*_{s;j}$.)

---

[3] We are not assuming that the Lagrange multipliers correspond to marginals; nonetheless, our choice of notation is deliberately suggestive, in that our proof shows that the Lagrange multipliers can be interpreted as local (pseudo)marginals.

In addition, each $\theta(\mathcal{T})$ is restricted to correspond to a tree-structured distribution, meaning that indices $\alpha \in \mathcal{I}$ corresponding to edges not in $\mathcal{T}$ must be zero. We let $\mathcal{I}(\mathcal{T}) \subset \mathcal{I}$ correspond to the set of exponential parameters that are free to vary for tree $\mathcal{T}$. (I.e., $\mathcal{I}(\mathcal{T}) = \mathcal{I}(V) \cup \mathcal{I}(E(\mathcal{T}))$).

Now the Lagrangian is also convex as a function of $\boldsymbol{\theta}$, so that it has a global minimum, attained at some $\widehat{\boldsymbol{\theta}} = \{\widehat{\theta}(\mathcal{T})\}$. By taking derivatives of the Lagrangian with respect to $\theta_\alpha$ for $\alpha \in \mathcal{I}(\mathcal{T})$ and using Lemma 1, we obtain the stationary conditions $\mu(\mathcal{T})\{\mathbb{E}_{\widehat{\theta}(\mathcal{T})}[\phi_\alpha(\mathbf{x}_\alpha)] - \widehat{T}_\alpha\} = 0$ for the optimum. If $\mu(\mathcal{T}) = 0$, then the tree parameter $\theta(\mathcal{T})$ plays no role in the problem, so that we can simply ignore it. Otherwise, if $\mu(\mathcal{T}) > 0$, for all indices $\alpha$, the Lagrange multipliers $\widehat{T}_\alpha$ are connected to the optimal tree parameters $\widehat{\theta}(\mathcal{T})$ via the relation:

$$\mathbb{E}_{\widehat{\theta}(\mathcal{T})}[\phi_\alpha(\mathbf{x}_\alpha)] = \widehat{T}_\alpha \qquad \text{for all } \alpha \in \mathcal{I}(\mathcal{T}) \qquad (10)$$

Recall that

$$\phi_\alpha(\mathbf{x}_\alpha) = \begin{cases} \delta(x_s = j) & \text{if } \alpha = (s; j) \\ \delta(x_s = j)\delta(x_t = k) & \text{if } \alpha = (st; jk) \end{cases}$$

so that the expectations $\mathbb{E}_{\widehat{\theta}(\mathcal{T})}[\phi_\alpha(\mathbf{x}_\alpha)]$ correspond to elements of the marginal probabilities; for example, we have $\mathbb{E}_{\widehat{\theta}(\mathcal{T})}[\phi_{s;j}(x_s)] = p(x_s = j; \widehat{\theta}(\mathcal{T}))$. As a consequence, equation (10) has two important implications:

(a) for all nodes $s \in V$, the single node marginals $p(x_s; \widehat{\theta}(\mathcal{T}))$ are all equal to a common quantity $\widehat{T}_s(x_s)$.

(b) similarly, for all trees $\mathcal{T}$ that include edge $(s,t)$, the joint marginal $p(x_s, x_t; \widehat{\theta}(\mathcal{T}))$ is equal to $\widehat{T}_{st}(x_s, x_t)$.

By the Legendre duality between the log partition function and the negative entropy function [see 14], we have the relation:

$$\Phi(\widehat{\theta}(\mathcal{T})) = \sum_{\alpha \in \mathcal{I}(\mathcal{T})} \widehat{\theta}(\mathcal{T})_\alpha \widehat{T}_\alpha - \Psi(\Pi^{\mathcal{T}}(\widehat{\mathbf{T}})) \qquad (11)$$

where $\Psi(\Pi^{\mathcal{T}}(\widehat{\mathbf{T}}))$ is the negative entropy of the tree-structured distribution $p(\mathbf{x}; \Pi^{\mathcal{T}}(\mathbf{T}))$. Substituting equation (11) into the definition of the Lagrangian yields an explicit expression for the negative of the Lagrangian dual function:

$$\mathcal{F}(\mathbf{T}; \vec{\mu}; \theta^*) = \mathbb{E}_{\vec{\mu}}[\Psi(\Pi^{\mathcal{T}}(\mathbf{T}))] - \sum_{\alpha \in \mathcal{I}} \mathbf{T}_\alpha \theta^*_\alpha \qquad (12)$$

We note that the following decomposition of the negative entropy for any tree-structured distribution:

$$\Psi(\Pi^{\mathcal{T}}(\widehat{\mathbf{T}})) = -\sum_{s \in V} H_s(T_s) + \sum_{(s,t) \in E(\mathcal{T})} I_{st}(T_{st})$$

Taking averages with respect to $\vec{\mu}$, we recover the form of $\mathcal{F}$ given in equation (8).

Since $\mathbf{T}$ must correspond to a set of pseudomarginals valid for each node and edge, it is restricted to the constraint set $\mathbb{C}$. Since the cost function is convex and the constraints are linear, strong duality holds [1]; therefore, the optimum dual[4] value $-\min_{\mathbf{T} \in \mathbb{C}} \mathcal{F}(\mathbf{T}; \vec{\mu}; \theta^*)$ is equivalent to the global minimum of the primal problem (4). $\square$

### 3.2 Optimization of the distribution $\vec{\mu}$

We now consider the problem of optimizing the choice of the distribution $\vec{\mu}$ over spanning trees. Since the function $\mathcal{F}$ only depends on this distribution via the *edge appearance probabilities* $\mu_e$, it is equivalent to optimize this vector, subject to the constraint that it belong to the spanning tree polytope $\mathbb{T}(G)$. We define the function $\mathcal{H}(\mu_e; \theta^*) \triangleq \min_{\mathbf{T} \in \mathbb{C}} \mathcal{F}(\mathbf{T}; \mu_e; \theta^*)$, where $\mu_e$ belongs to $\mathbb{T}(G)$.

**Theorem 2 (Jointly optimal bounds).**
*We have an upper bound, jointly optimal over both $\mathbf{T}$ and $\mu_e$, of the form:*

$$\begin{aligned} \Phi(\theta^*) &\leq - \max_{\mu_e \in \mathbb{T}(G)} \mathcal{H}(\mu_e; \theta^*) \\ &= - \max_{\mu_e \in \mathbb{T}(G)} \min_{\mathbf{T} \in \mathbb{C}} \mathcal{F}(\mathbf{T}; \mu_e; \theta^*) \qquad (13) \end{aligned}$$

*The function $\mathcal{H}$ has a global maximum attained by some $\widehat{\mu_e}$, which yields the tightest possible upper bound.*

**Remarks:** (1) Let $\widehat{\vec{\mu}}$ be a distribution with edge appearance probabilities $\widehat{\mu_e}$. Then it can be shown [14] that the optimal pair $(\widehat{\mu_e}; \widehat{\mathbf{T}}(\widehat{\mu_e}))$ satisfies the relations:

$$\sum_{e \in E(\mathcal{T})} I_e(\widehat{\mathbf{T}}(\widehat{\mu_e})) = \sum_{e \in E} \widehat{\mu}_e I_e(\widehat{\mathbf{T}}(\widehat{\mu_e}))$$

for trees $\mathcal{T} \in \text{supp}(\widehat{\vec{\mu}})$. This condition corresponds to a type of equalization of mutual information on trees.

(2) Moreover, it can be shown [14] that the following minimax relation holds:

$$\max_{\mu_e \in \mathbb{T}(G)} \min_{\mathbf{T} \in \mathbb{C}} \mathcal{F}(\mathbf{T}; \mu_e; \theta^*) = \min_{\mathbf{T} \in \mathbb{C}} \max_{\mu_e \in \mathbb{T}(G)} \mathcal{F}(\mathbf{T}; \mu_e; \theta^*)$$

This relation has a game-theoretic interpretation: we can think of a two-person game, in which Player 1 chooses local pseudomarginals $\mathbf{T}$ so as to minimize $\mathcal{F}(\mathbf{T}; \mu_e; \theta^*)$, whereas Player 2 chooses a spanning tree so as to maximize the same quantity. The minimax relation specifies that a *mixed strategy* is optimal, in the sense that Player 2 chooses not just a single spanning tree, but rather a probability distribution over spanning trees.

---

[4] We have defined $\mathcal{F}$ as the *negative* of the Lagrangian dual, so that it should be minimized as opposed to maximized.

*Proof of Theorem 2*: The bound of equation (9) holds for all $\mu_e \in \mathbb{T}(G)$, from which equation (13) follows. Observe that $\mathcal{F}(\lambda; \mu_e; \theta^*)$ is linear in $\mu_e$. Therefore, $\mathcal{H}(\mu_e; \theta^*)$ is the minimum of a collection of linear functions, and so is concave as a function of $\mu_e$ [1]. Consequently, $\mathcal{H}(\mu_e; \theta^*)$ has a globally optimal point $\widehat{\mu_e}$, at which the optimal value of the upper bound in equation (13) is attained. □

We illustrate Theorems 1 and 2 by following up the single cycle case of Example 1.

**Example 2 (Optima on a single cycle).**
Consider the single cycle of Example 1, but the non-symmetric choice of exponential parameter $\theta^* = [0\ 0\ 0\ 0\ 1\ 1\ 3]^T$. If we choose uniform (3/4) edge appearance probabilities, then we obtain an upper bound $-\mathcal{F}(\widehat{\mathbf{T}}; 3/4; \theta^*) \approx 6.3451$, optimal in the sense of Theorem 1, on the log partition function $\Phi(\theta^*) \approx 6.3326$. If we perform the optimization of the edge weights $\mu_e$ (by an algorithm to be specified), we obtain the optimal edge appearance probabilities $\widehat{\mu_e} \approx [0.92\ 0.54\ 0.54\ 1]$. Note that the optimum assigns edge appearance probability of one to the edge with largest weight (i.e., the single edge with weight 3). As a result, this edge must appear in any spanning tree in the support of an optimal distribution $\widehat{\overline{\mu}}$. This set of edge appearance probabilities, combined with the associated $\widehat{\mathbf{T}}(\widehat{\mu_e})$, yields the upper bound $-\mathcal{H}(\widehat{\mu_e}; \theta^*) \approx 6.3387$ on the true log partition function $\Phi(\theta^*) \approx 6.3326$. This upper bound is tighter than the previous bound ($\approx 6.3451$) based on uniform edge appearance probabilities.

## 4 Algorithms for optimization

We first consider the problem of computing the upper bound of Theorem 1 (that is, with $\mu_e \in \mathbb{T}(G)$ fixed). This task requires minimizing the function $\mathcal{F}(\mathbf{T}; \mu_e; \theta^*)$ defined in equation (8), subject to the constraint that $\mathbf{T} \in \mathbb{C}$. Since the problem is convex with linear constraints, a variety of methods can be used. In our current work, we have used a form of constrained Newton's method [1], which has desirable convergence properties. In future work, we will describe a modified form of local message-passing, analogous to but distinct from belief propagation, for solving this variational problem.

Next we consider the problem of maximizing $\mathcal{H}$ over $\mu_e \in \mathbb{T}(G)$, as required for the upper bound of Theorem 2. Since neither the Hessian nor the gradient of $\mathcal{H}$ are difficult to compute, it is tempting to apply a constrained Newton's method once again. However, the spanning tree polytope $\mathbb{T}(G)$ is defined [see 14] by a very large number of linear inequalities ($\mathcal{O}(2^N)$), which precludes solving the quadratic programs required by constrained Newton's method.

Interestingly, it turns out that despite the exponential number of constraints characterizing $\mathbb{T}(G)$, maximizing a linear function over this polytope is feasible. Indeed, this task is equivalent to solving a *maximum weight spanning tree* problem [see 8]. On this basis, it can be seen that the conditional gradient method [1], as specified in Algorithm 1, is a computationally feasible proposal.

---

**Algorithm 1 (Conditional gradient).**

1. Initialize $\mu_e^0 \in \mathbb{T}(G)$.

2. For iterations $n = 0, 1, 2, \ldots$, compute the ascent direction as follows:
$$\widetilde{\mu_e}^{n+1} = \arg \max_{\mu_e \in \mathbb{T}(G)} \left\{ \langle \nabla \mathcal{H}(\mu_e^n; \theta^*), (\mu_e - \mu_e^n) \rangle \right\}$$

3. Form $\mu_e^{n+1} = (1 - \alpha^n)\mu_e^n + \alpha^n \widetilde{\mu_e}^{n+1}$ where $\alpha^n \in (0, 1)$ is a step size parameter.

---

It can be shown [14] that elements of the gradient $\nabla \mathcal{H}(\mu_e^n; \theta^*)$ correspond to mutual information terms $I_{st}(\widehat{\mathbf{T}}_{st}(\mu_e^n))$. As a consequence, the second and third steps of Algorithm 1 have an interesting interpretation. In particular, let us view the current pseudo-marginal vector $\widehat{\mathbf{T}}(\mu_e^n)$ as a set of data, which is used to specify mutual information terms $I_{st}(\widehat{T}_{st}(\mu_e^n))$ on each edge. In this case, finding the corresponding maximum weight spanning tree is equivalent to finding the tree distribution that best fits the data in the maximum likelihood sense, or Kullback-Leibler divergence between the empirical distribution specified by the data, and the tree distribution. (See Chow and Liu [3] for more details on this interpretation of the maximum weight spanning tree procedure.) Consequently, at each iteration, the algorithm takes a step towards the spanning tree that best fits the current data. The size $\alpha^n$ of this step is chosen by the limited minimization or Armijo rule [see 1].

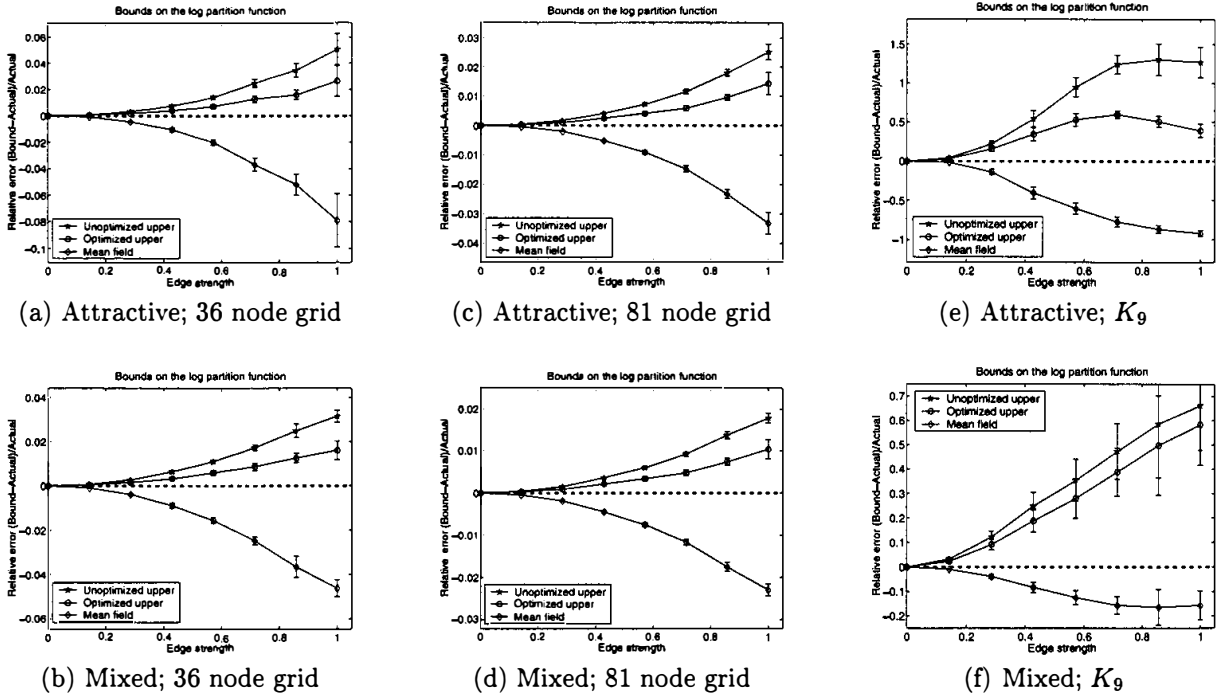

Figure 2. Upper and lower bounds on $\Phi(\theta^*)$ for a randomly chosen distribution $p(\mathbf{x}; \theta^*)$ on various graphs: 36 node grid (first column), 81 node grid (middle column) or fully connected 9 node graph $K_9$ (right column). Panels in the top (respectively bottom ) row correspond to the attractive (respectively mixed) condition. Each panel shows the mean relative error $[\text{Bound} - \Phi(\theta^*)]/\Phi(\theta^*)$ versus a normalized measure of edge strength; error bars correspond to plus/minus one standard deviation. See text for more details.

## 5 Results

To illustrate our bounds, we present the results of simulations on three graphs: square 2-D grids with 36 or 81 nodes, as well as a fully connected graph with 9 nodes ($K_9$). We performed simulations for a binary process, using the standard minimal exponential representation of $\theta^* = \{\theta_s^*, \theta_{st}^*\}$ of the Ising model.[5] In all cases, we set $\theta_s^* = 0$ for all $s \in V$. For a given edge strength $d > 0$, we set the pairwise potentials in one of two ways: (a) for an *attractive* ensemble, choose $\theta_{st}^* \sim \mathcal{U}[0, d]$ independently for each edge; (b) for a *mixed* ensemble, choose $\theta_{st}^* \sim \mathcal{U}[-d, d]$ independently for each edge.[6] For each of graphs and each of the two conditions (attractive or mixed), we ran simulations with edge strengths $d$ ranging from 0 to $\frac{4}{\sqrt{N}}$. For each condition and setting of the edge strength, we performed 30 trials for the grids, and 10 trials for $K_9$. The inner optimization $\min_{\mathbf{T} \in \mathcal{C}} \mathcal{F}(\mathbf{T}; \boldsymbol{\mu}_e; \theta^*)$ was performed using constrained Newton's method [1], whereas the outer maximization was performed with the conditional gradient method (Algorithm 1). In all cases, step size choices were made by the Armijo rule [1]. The value of the actual partition function $\Phi(\theta^*)$ was computed by forming a junction tree for each grid, and performing exact computations on this junction tree.

Shown in Figure 2 are plots of the *relative error* $[\text{Bound} - \Phi(\theta^*)]/\Phi(\theta^*)$ versus the edge strength (normalized by $4/\sqrt{N}$ for each $N$ so that it falls in the interval $[0, 1]$). The "unoptimized" curve shows the bound of Theorem 1 with the fixed choice of uniform edge appearance probabilities $\mu_e = (N-1)/|E|$. The "optimized" curve corresponds to the jointly optimal (over both $\mathbf{T}$ and $\vec{\mu}$) upper bounds of Theorem 2. The lower curve in each panel corresponds to the relative error in the naive mean field lower bound. Note that the gain from optimizing $\boldsymbol{\mu}_e$ is especially pronounced as the edge strength is increased, in which case the distribution of edge weights $\theta_{st}^*$ becomes more inhomogeneous. For the two square grids, the tightness of the upper bounds decreases more slowly than the corresponding mean field lower bound. In terms of the relative error plotted here, the upper bounds are superior to the mean field bound by factors of roughly 3 and 2 in the attractive and mixed cases respectively. For the fully connected $K_9$, all of the bounds are much poorer. In the attractive condition, the fully optimized upper bound remains superior to the mean field bound. In the mixed condition, the mean field lower bound is of mediocre quality, whereas the upper bounds are very poor. Thus, the quality of our upper bounds appears to degrade for mixed potentials on densely connected graphs.

---

[5]For these simulations, we used a set of so-called spin variables taking values in $\{-1, +1\}$.

[6]The notation $\mathcal{U}[a, b]$ denotes the uniform distribution on $[a, b]$.

It is worthwhile emphasizing the importance of the dual formulation of our bounds. Indeed, the naive approach of attempting to optimize the primal problem (4) would require dealing with a number[7] of spanning trees that ranges from $4,782,969$ for $K_9$, all the way up to the astronomical number $\approx 8.33 \times 10^{33}$ for the grid with $N = 81$ nodes.

## 6 Conclusions

We have developed and analyzed a new class of upper bounds for the log partition function that are based on convex combinations of the exponential parameters corresponding to a set of tractable distributions. This paper treated in detail the case of convex combinations of spanning trees. Despite the prohibitively large number of spanning trees in a general graph, we developed a technique for optimizing the bounds efficiently — though implicitly — over all spanning trees.

Although this paper focused on the special case of graphs with pairwise cliques, the line of analysis outlined here is broadly applicable. For instance, it is natural to consider convex combinations of more complex approximating structures — for example, hypertrees of width $k \geq 2$, as opposed to spanning trees.[8] Doing so leads to dual functions that correspond to "convexified" forms of Kikuchi and related free energies. Optimization of the pseudomarginals $\mathbf{T}$, which now involve higher order cliques, is again possible. Since these families of substructures are nested (e.g., spanning trees are a strict subset of graphs of treewidth 2), this procedure provides a sequence of progressively tighter bounds. However, there is a caveat: unlike the spanning tree case, the optimization of the distribution $\vec{\mu}$ over hypertrees is not straightforward. Although solving the maximum weight spanning tree problem required as part of Algorithm 1 is easy, its analog for hypertrees of width $k \geq 2$ is NP-hard [12].

The results of this paper, in addition to the usefulness of the upper bounds, have possible implications for approximate inference. As with belief propagation and the Bethe free energy [15], the variational problems specified by Theorem 1 suggest the following agenda: after performing the minimization, take the optimizing argument $\widehat{\mathbf{T}}$ as an approximation to the marginals of $p(\mathbf{x}; \theta^*)$. One possible advantage of the variational problems in Theorem 1 is that, in contrast with the Bethe free energy, they have a unique global minimum that can be found by a variety of techniques.

### Acknowledgements

Work supported by ODDR&E MURI Grant DAAD19-00-1-0466 through ARO; by ONR Grant N00014-00-1-0089; and by AFOSR Grant F49620-00-1-0362. Thanks to Yee Whye Teh and Max Welling for sharing their code for performing exact inference on grids. Thanks to David Karger for pointers to the spanning tree polytope.

---

[7] These numbers can be calculated by applying the Matrix-Tree theorem [2].

[8] Background on hypertrees and hypergraphs can be found in the thesis [12].